\documentclass[11pt]{article}

\usepackage{acl}
\usepackage{times}
\usepackage{latexsym}
\usepackage[T1]{fontenc}
\usepackage[utf8]{inputenc}
\usepackage{microtype}
\usepackage{inconsolata}
\usepackage{graphicx}
\usepackage{booktabs}
\usepackage{amsmath}
\usepackage{amssymb}
\usepackage{multirow}
\usepackage{xcolor}
\usepackage[skins,breakable]{tcolorbox}

\definecolor{bestgreen}{RGB}{0,128,0}

\newtcolorbox{appendixbox}[1]{
  breakable,
  colback=black!1,
  colframe=black!35,
  boxrule=0.6pt,
  arc=2pt,
  left=6pt,
  right=6pt,
  top=4pt,
  bottom=4pt,
  before upper=\raggedright,
  title=\textbf{#1}
}

\begin{document}

\newpage
\twocolumn
\pagestyle{plain}
\setcounter{page}{1}

\title{Attention Isn't All You Need for Emotion Recognition: \\ Domain Features Outperform Transformers on the EAV Dataset}

\author{Anmol Guragain \\
  Universidad Politécnica de Madrid \\
  E.T.S. Ingenieros de Telecomunicación \\
  \texttt{anmol.g@upm.es}}

\maketitle

\begin{abstract}
We present a systematic study of multimodal emotion recognition using the EAV dataset, investigating whether complex attention mechanisms improve performance on small datasets. We implement three model categories: baseline transformers (M1), novel factorized attention mechanisms (M2), and improved CNN baselines (M3). Our experiments show that sophisticated attention mechanisms consistently underperform on small datasets. M2 models achieved 5 to 13 percentage points below baselines due to overfitting and destruction of pretrained features. In contrast, simple domain-appropriate modifications proved effective: adding delta MFCCs to the audio CNN improved accuracy from 61.9\% to \textbf{65.56\%} (+3.66pp), while frequency-domain features for EEG achieved \textbf{67.62\%} (+7.62pp over the paper baseline). Our vision transformer baseline (M1) reached \textbf{75.30\%}, exceeding the paper's ViViT result (74.5\%) through domain-specific pretraining, and vision delta features achieved \textbf{72.68\%} (+1.28pp over the paper CNN). These findings demonstrate that for small-scale emotion recognition, domain knowledge and proper implementation outperform architectural complexity.
\end{abstract}

\section{Introduction}

Emotion recognition from physiological and behavioral signals is a core challenge in affective computing, with applications in human-computer interaction, mental health monitoring, and assistive technologies. The EAV dataset \cite{eav2024} provides synchronized EEG, Audio, and Video recordings from 42 participants engaged in conversational interactions, offering a unique resource for multimodal emotion research alongside large-scale multimodal affect datasets such as CMU-MOSEI \cite{zadeh2018mosei}. However, with only approximately 280 training samples per subject after data splitting, this dataset presents significant challenges for deep learning approaches.

This limited data availability motivates our central research question: \textit{Do complex attention mechanisms improve emotion recognition performance on small datasets, or do simpler approaches with domain-appropriate features perform better?}

To answer this question, we systematically evaluate three categories of models. First, we implement baseline transformer architectures (M1) to establish reference performance. Second, we develop novel factorized attention mechanisms (M2) tailored to each modality's unique structure: spatial-temporal-asymmetry attention for EEG, temporal-frequency dual attention for audio, and ViViT-style space-time attention for video. Third, we explore minimal CNN improvements (M3) focusing on bug fixes and domain-appropriate feature engineering.

Our study makes three contributions: (1) we show that factorized attention mechanisms consistently fail on small datasets, underperforming baselines by 5 to 13 percentage points; (2) we demonstrate that simple, domain-appropriate modifications (delta MFCCs for audio and band power features for EEG) achieve state-of-the-art results on this dataset; and (3) we provide analysis of failure modes and success factors for emotion recognition on limited data.

\section{Related Work}

\subsection{Emotion Recognition Datasets}

Traditional EEG datasets such as DEAP \cite{deap2012} and MAHNOB-HCI \cite{mahnob2012} employ passive elicitation paradigms where participants watch emotional stimuli. Audio-visual datasets like IEMOCAP \cite{iemocap2008} capture conversational emotions but lack neurophysiological signals. The EAV dataset bridges this gap by providing synchronized multimodal recordings during active conversation, though its relatively small scale (42 participants) constrains the complexity of applicable models.

\subsection{Transformer Architectures for Emotion Recognition}

The Audio Spectrogram Transformer (AST) \cite{gong2021ast} treats spectrograms as images, applying Vision Transformer architecture with strong performance on large-scale audio datasets. Video Vision Transformer (ViViT) \cite{arnab2021vivit} extends this approach to video through factorized space-time attention, achieving 22$\times$ computational efficiency over full attention. EEGformer \cite{eegformer} adapts transformers for EEG signals. These architectures achieve impressive results on large datasets, but their effectiveness on small-scale emotion recognition remains unclear.

\subsection{Frontal Asymmetry in Emotion}

Davidson's frontal asymmetry model \cite{davidson1992} establishes that left frontal cortex activation (measured at F3) correlates with approach motivation and positive emotions, while right frontal activation (F4) correlates with withdrawal motivation and negative emotions. This neurophysiological insight motivates our M2 EEG architecture and M3 band power features.

\subsection{Feature Engineering for Emotion Recognition}

Classical approaches to emotion recognition rely on hand-crafted features. For EEG, band power and differential entropy features computed from frequency bands (delta, theta, alpha, beta, gamma) have proven effective on the SEED dataset \cite{zheng2015seed} and remain widely used across EEG emotion recognition studies \cite{patel2021entropy}. For audio, Mel-frequency cepstral coefficients (MFCCs) and their temporal derivatives (delta and delta-delta) capture both spectral content and dynamics \cite{schuller2013interspeech,elayadi2011survey,ververidis2006emotional}, and are commonly extracted using toolkits such as openSMILE \cite{eyben2010opensmile}. Squeeze-and-Excitation networks \cite{hu2018senet} introduced channel attention for CNNs, learning to weight feature map importance through a bottleneck architecture.

\section{Dataset}

The EAV dataset contains recordings from 42 participants, each providing 200 interaction samples across five emotion classes: Neutral, Anger, Happiness, Sadness, and Calmness. EEG signals are recorded using a 30-channel BrainAmp system at 500 Hz, audio at 16 kHz, and video at 30 fps.

Following standard preprocessing, EEG signals are downsampled to 100 Hz and bandpass filtered (0.5 to 45 Hz), yielding segments of shape (30 channels, 500 samples). Audio clips of 5 seconds are extracted, and 25 video frames are sampled per segment. The data is split 70/30 for training and testing per subject, resulting in approximately 280 training and 120 test samples per subject per modality.

This limited training data (280 samples with 15,000 EEG features, 80,000 audio samples, or 25$\times$224$\times$224$\times$3 video pixels) fundamentally constrains model complexity and motivates our investigation of simpler approaches.

\section{Methods}

\subsection{M1: Baseline Transformer Architectures}

Our baseline models replicate the paper's transformer approach using pretrained architectures from Hugging Face.

\paragraph{EEG Transformer.} We implement a custom architecture with convolutional feature extraction followed by transformer encoding. Two 1D convolutional layers (kernel size 11, padding 5) transform the 30-channel input into 60-dimensional features while preserving temporal resolution. Six transformer encoder layers with 4 attention heads and hidden dimension 60 model temporal dependencies. Mean pooling followed by a two-layer classifier produces emotion predictions. This architecture applies standard self-attention across time, treating channels as input features rather than a separate dimension.

\paragraph{Audio Spectrogram Transformer.} We fine-tune the pretrained AST model (\texttt{MIT/ast-finetuned-audioset}) with a custom 5-class classifier head. The model converts audio waveforms to mel spectrograms (1024$\times$128), extracts 16$\times$16 patches yielding approximately 1212 tokens, and applies 12 transformer layers with 768-dimensional embeddings. Training proceeds in two stages: 10 epochs with frozen backbone (learning rate 5e-4), followed by 15 epochs with unfrozen backbone (learning rate 5e-6).

\paragraph{Vision Transformer.} We employ a ViT model pretrained on facial emotion recognition (\texttt{dima806/facial\_emotions\_image\_detection}). Each video is processed as 25 independent frames, with per-frame predictions averaged for the final classification. The same two-stage training strategy is applied: 10 epochs frozen, then 5 epochs unfrozen. While this approach ignores temporal dynamics across frames, the facial emotion pretraining (often built from large-scale facial affect data such as AffectNet) provides strong baseline performance \cite{mollahosseini2017affectnet}.

\subsection{M2: Factorized Attention Mechanisms}

\begin{figure*}[t]
\centering
\includegraphics[width=\textwidth]{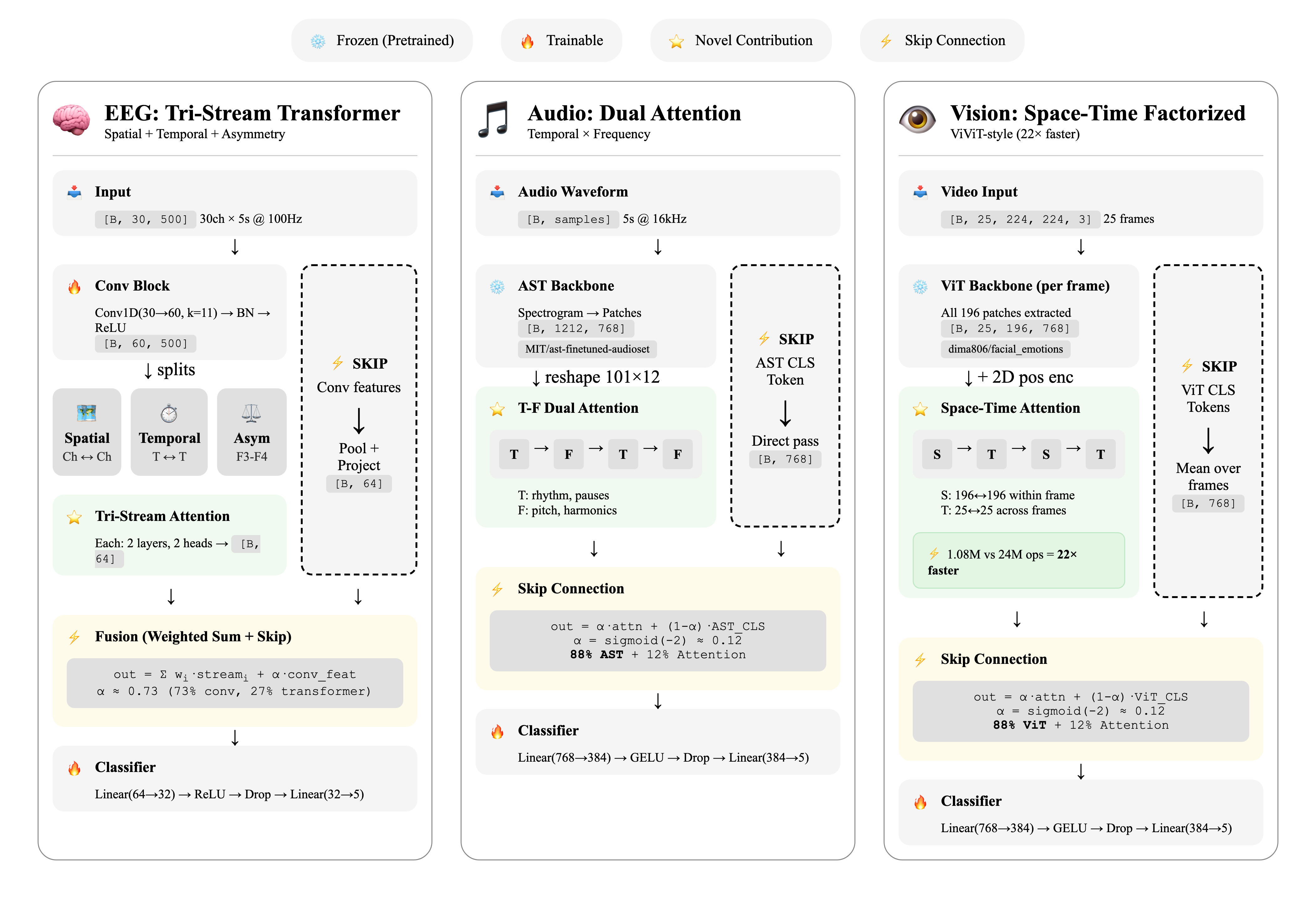}
\caption{Architecture diagrams for M2: Factorized Attention Mechanisms. Left: EEG Tri-Stream Transformer with spatial, temporal, and asymmetry attention streams. Center: Audio Dual Attention with temporal-frequency factorization over AST features. Right: Vision Space-Time Attention with factorized spatial and temporal attention over ViT embeddings. All architectures include skip connections to preserve pretrained features.}
\label{fig:m2_architecture}
\end{figure*}

Our main architectural contribution explores whether factorized attention, tailored to each modality's structure, can improve upon baselines. Figure~\ref{fig:m2_architecture} illustrates the three M2 architectures.

\subsubsection{M2-EEG: Tri-Stream Transformer}

EEG signals contain information along three distinct axes. The spatial dimension captures relationships between brain regions (channels), the temporal dimension captures signal dynamics, and the hemispheric asymmetry dimension captures left-right differences relevant for emotional valence.

Our architecture first applies convolutional feature extraction, then branches into three parallel attention streams. The spatial stream transposes features to (time, channels, features) and applies attention across channels, learning which brain regions interact at each time point. The temporal stream applies attention across time points for each channel, capturing event-related dynamics. The asymmetry stream computes difference signals between six hemisphere pairs (Fp1-Fp2, F3-F4, F7-F8, C3-C4, P3-P4, O1-O2) based on the 10-20 electrode system, then applies attention across these pairs. Following Davidson's model, F3-F4 receives a 1.2$\times$ attention weight.

Stream outputs are fused using learned softmax weights, with a learnable skip connection preserving convolutional features. The skip weight is initialized such that approximately 73\% of the signal comes from the convolutional path. We explored three model sizes (138K, 419K, and 1.6M parameters), with the smallest recommended for this dataset size.

\subsubsection{M2-EEG: EEGNet-Improved}

As a literature-based alternative, we enhanced EEGNet with techniques from recent papers: squeeze-excitation blocks for channel attention, multi-scale temporal convolution (kernels 16, 32, 64), residual connections, data augmentation (amplitude scaling, Gaussian noise, time shift), label smoothing (0.1), AdamW optimizer with weight decay, and cosine annealing learning rate schedule.

\subsubsection{M2-Audio: Temporal-Frequency Dual Attention}

Audio spectrograms have natural 2D structure: the temporal axis captures speech rhythm, pauses, and intensity changes, while the frequency axis captures pitch, harmonics, and voice quality. Standard AST flattens this structure into a 1D patch sequence, losing the distinction.

Our architecture preserves AST's pretrained features but adds factorized attention. After extracting 1212 patches from AST, we reshape them into a 101$\times$12 grid (time$\times$frequency). Learnable 2D positional encodings are added. Two dual-attention blocks apply temporal attention (each frequency band attends across time) followed by frequency attention (each time step attends across frequencies). A learnable skip connection, initialized with weight $-2.0$ (yielding $\alpha \approx 0.12$), blends the attention output with AST's original CLS token, preserving approximately 88\% of pretrained features.

\subsubsection{M2-Vision: Factorized Space-Time Attention}

Video understanding requires both spatial analysis within frames and temporal modeling across frames. Full space-time attention over 25 frames $\times$ 196 patches has prohibitive $O((25 \times 196)^2) \approx 24$ million attention entries.

Following ViViT's Model 2 (factorized encoder), we decompose this into spatial attention within each frame ($O(25 \times 196^2)$) followed by temporal attention for each patch position across frames ($O(196 \times 25^2)$), achieving 22$\times$ speedup. Four factorized attention blocks process ViT's patch embeddings (shape: batch$\times$25$\times$196$\times$768), with separate learnable spatial and temporal positional encodings. A skip connection preserves ViT's pretrained features.

\subsection{M3: CNN Improvements}

After M2's poor results, we investigated whether simple fixes to baseline CNNs could outperform complex attention mechanisms.

\subsubsection{M3-Audio: Delta MFCCs}

The original audio CNN uses 180 features (40 MFCCs, 12 chroma, 128 mel), averaged over time. To introduce temporal dynamics with minimal additional capacity, we augment the feature set with delta MFCCs. Delta MFCCs approximate the first temporal derivative of the cepstral coefficients and capture short-term spectral change patterns that are informative for affective speech (e.g., variations in pitch and energy).

Our modification adds 40 delta MFCC features (also time-averaged), increasing the feature vector from 180 to 220 dimensions. Combined with AdamW optimizer (weight decay 1e-4) and label smoothing (0.1), this minimal change achieved our best audio result.

\subsubsection{M3-EEG v1: Bug Fixes}

We identified an implementation issue in the original EEGNet classifier: applying \texttt{nn.Softmax(dim=1)} before \texttt{CrossEntropyLoss}, which already applies (log-)softmax internally. This double normalization can distort gradients and hinder optimization. We therefore removed the explicit softmax and trained the model on raw logits, using AdamW with weight decay (0.01) and label smoothing (0.1).

\subsubsection{M3-EEG v2: Band Power Features}

Bug fixes alone proved insufficient, as raw time-domain EEG is often heavily contaminated by artifacts (e.g., muscle activity, eye blinks, and environmental noise). Following established practice in EEG emotion recognition (e.g., SEED-style pipelines) \cite{zheng2015seed}, we therefore extracted frequency-domain features.

Band power features compute power spectral density using Welch's method for five frequency bands (delta: 0.5 to 4 Hz, theta: 4 to 8 Hz, alpha: 8 to 13 Hz, beta: 13 to 30 Hz, gamma: 30 to 45 Hz) across all 30 channels, yielding 150 features. Differential entropy features compute $\text{DE} = 0.5 \log(2\pi e \sigma^2)$ for each band-filtered signal, adding another 150 features. Alpha asymmetry features compute $\ln(P_{\text{right}}^\alpha) - \ln(P_{\text{left}}^\alpha)$ for six hemisphere pairs, adding 6 features that capture frontal asymmetry.

The 306-dimensional feature vector feeds a simple MLP (306$\rightarrow$128$\rightarrow$64$\rightarrow$5) with batch normalization and 50\% dropout. Despite having more parameters than EEGNet, this approach operates on meaningful, low-noise features rather than raw high-dimensional signals.

\paragraph{Illustrative Comparison.} The raw EEG segment contains 15,000 samples (30 channels $\times$ 500 time points) and includes substantial nuisance variability. In contrast, the 306-dimensional feature representation applies (i) a frequency-domain transformation with averaging (Welch), (ii) band selection that emphasizes physiologically meaningful rhythms (e.g., alpha/beta), and (iii) explicit hemispheric asymmetry features. Learning an equivalent denoising and summarization pipeline end-to-end from $\sim$280 training samples per subject is unlikely to generalize.

\subsubsection{M3-Vision v1: SE Block Fix}

We found that the squeeze-excitation (SE) module in the original video CNN used reduction ratio 1, i.e., no bottleneck (2048$\rightarrow$2048$\rightarrow$2048 rather than 2048$\rightarrow$128$\rightarrow$2048). Since SE relies on a bottleneck to learn channel reweighting efficiently \cite{hu2018senet}, we set the reduction ratio to 16.

\subsubsection{M3-Vision v2: Delta Features}

Following our audio success, we added temporal delta features to the video CNN, capturing frame-to-frame changes in facial expression. Delta features compute the difference between consecutive frame representations, encoding expression dynamics that static features miss.

\begin{figure*}[t]
\centering
\includegraphics[width=\textwidth]{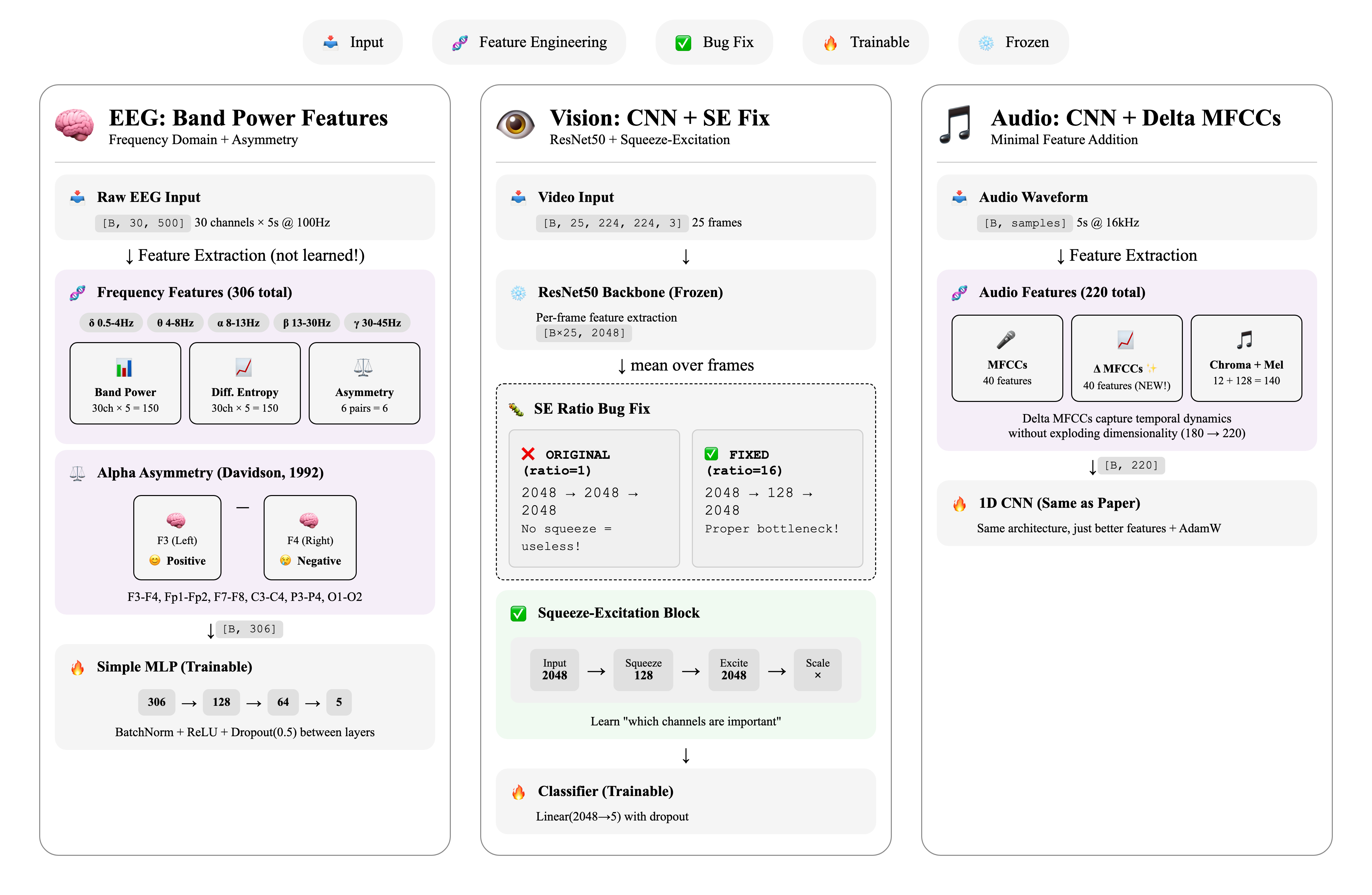}
\caption{Architecture diagrams for M3: CNN Improvements. Left: EEG Band Power Features with frequency-domain feature extraction (band power, differential entropy, and alpha asymmetry) feeding a simple MLP. Center: Vision CNN with ResNet50 backbone and fixed squeeze-excitation block. Right: Audio CNN with delta MFCC features added to the original feature set.}
\label{fig:m3_architecture}
\end{figure*}

\section{Experiments and Results}

All models were trained on NVIDIA RTX 6000 Ada GPUs with 48GB of VRAM. We report mean accuracy and standard deviation across all 42 subjects.

\begin{table*}[t]
\centering
\small
\begin{tabular}{@{}llccc@{}}
\toprule
\textbf{Category} & \textbf{Model} & \textbf{EEG} & \textbf{Audio} & \textbf{Vision} \\
\midrule
\multirow{2}{*}{Paper Baselines} 
& CNN & 60.00\% & 61.90\% & 71.40\% \\
& Transformer & 53.50\% & 62.70\% & 74.50\% \\
\midrule
M1: Our Baseline & Transformer & 52.68\% $\pm$ 10.3 & 58.06\% $\pm$ 10.5 & \textcolor{bestgreen}{\textbf{75.30\%}} $\pm$ 10.5 \\
\midrule
\multirow{4}{*}{M2: Factorized Attn.}
& EEG Tri-Stream & 48.00\% $\pm$ 12.1 & --- & --- \\
& EEGNet-Improved & 58.47\% $\pm$ 11.2 & --- & --- \\
& Audio Dual-Attn. & --- & 49.46\% $\pm$ 15.8 & --- \\
& Vision ViViT-style & --- & --- & 69.54\% $\pm$ 12.3 \\
\midrule
\multirow{5}{*}{M3: CNN Improved}
& EEG v1 (Bug Fix) & 57.66\% $\pm$ 13.7 & --- & --- \\
& EEG v2 (Band Power) & \textcolor{bestgreen}{\textbf{67.62\%}} $\pm$ 8.2 & --- & --- \\
& Audio (Delta MFCC) & --- & \textcolor{bestgreen}{\textbf{65.56\%}} $\pm$ 9.3 & --- \\
& Vision v1 (SE Fix) & --- & --- & 69.42\% $\pm$ 11.9 \\
& Vision v2 (Delta) & --- & --- & \textcolor{bestgreen}{\textbf{72.68\%}} $\pm$ 10.8 \\
\bottomrule
\end{tabular}
\caption{Results across all models. \textcolor{bestgreen}{\textbf{Green}} indicates results exceeding paper baselines. M2 factorized attention underperforms across all modalities; M3 simple fixes achieve best results.}
\label{tab:results}
\end{table*}

\begin{table}[t]
\centering
\small
\begin{tabular}{@{}lccc@{}}
\toprule
\textbf{Subject} & \textbf{EEG} & \textbf{Audio} & \textbf{Vision} \\
\midrule
\multicolumn{4}{l}{\textit{Top Performers}} \\
S17 & 82.50\% & 85.00\% & 93.33\% \\
S6 & --- & 78.33\% & 94.17\% \\
S20 & 83.33\% & --- & 91.67\% \\
S24 & 85.00\% & --- & --- \\
S2 & 80.83\% & 85.00\% & --- \\
\midrule
\multicolumn{4}{l}{\textit{Difficult Subjects}} \\
S35 & 46.67\% & 55.00\% & 56.67\% \\
S12 & --- & 54.17\% & 43.33\% \\
S29 & 56.67\% & 47.50\% & --- \\
S37 & --- & 46.67\% & --- \\
S34 & 54.17\% & 49.17\% & --- \\
\bottomrule
\end{tabular}
\caption{Subject-level analysis showing best M3 results. S17 is consistently the most expressive subject across all modalities.}
\label{tab:subjects}
\end{table}

Table~\ref{tab:results} summarizes results across all models. Three findings emerge.

\paragraph{M2 Factorized Attention Underperforms.} All M2 models fall below their respective baselines. The EEG Tri-Stream reaches 48\% (vs. a 60\% paper baseline), Audio Dual-Attention reaches 49.46\% (with several subjects near chance-level performance), and the ViViT-style video model reaches 69.54\% (vs. 74.5\%). Although the factorized designs are well motivated, they appear difficult to fit reliably in the low-sample regime (\,$\sim$280 training samples per subject).

\paragraph{M3 Simple Fixes Improve Performance.} Domain-appropriate feature design yields consistent gains. EEG band-power features reach 67.62\% (+7.62pp over the paper baseline), Audio delta MFCCs reach 65.56\% (+3.66pp), and Vision delta features reach 72.68\% (+1.28pp over the paper CNN). These results suggest that, under limited supervision, established signal-processing priors (frequency analysis and temporal derivatives) can provide more reliable inductive bias than additional attention modules. Figure~\ref{fig:m3_architecture} summarizes these models.

\paragraph{Vision: Pretraining Domain Matters.} For Vision, M1's domain-specific pretraining (75.30\%) outperformed all M3 CNN modifications (72.68\%), demonstrating that pretraining domain matters more than architectural changes. The ViT model pretrained on facial emotions provides better features than ResNet50 pretrained on ImageNet, even with our improvements.

\paragraph{Subject Variability.} Table~\ref{tab:subjects} shows substantial individual differences. Subject 17 consistently achieves the highest accuracy across all modalities (93.33\% vision, 85\% audio, 82.5\% EEG), suggesting relatively expressive emotional displays. Subjects 35 and 12 present persistent challenges, potentially due to subtle expressions or individual physiological differences.

\section{Analysis: Why M2 Underperformed}

Our M2 architectures likely underperformed for three interconnected reasons.

\paragraph{Degradation of Pretrained Representations.} AST and ViT backbones encode strong representations learned from large-scale pretraining. Although we include skip connections, the additional attention layers still perturb these features. The learned skip weight ($\alpha \approx 0.12$) indicates a strong preference for the pretrained pathway (\,$\approx$88\%), yet the residual attention contribution can still negatively affect downstream accuracy.

\paragraph{Insufficient Training Data.} The EEG Tri-Stream model has 138,401 parameters but only $\sim$280 training samples per subject. In this setting, model capacity can easily exceed what the data can constrain, leading to high variance and unstable generalization. Empirically, we observed strong overfitting (near-perfect training accuracy with substantially lower test accuracy), particularly for the M2 audio variant where multiple subjects collapsed toward chance-level performance.

\paragraph{Complex Inductive Biases Require Data.} Factorized attention introduces additional structure (e.g., separate spatial, temporal, and frequency reasoning). However, exploiting these inductive biases still requires sufficient supervision to estimate the extra parameters and to learn stable fusion. With limited data, the added flexibility can dominate any benefit and reduce robustness.

\section{Analysis: Why M3 Succeeded}

\paragraph{Domain Knowledge Over Learned Features.} Band power features incorporate well-established neurophysiological priors into a compact 306-dimensional representation. Alpha asymmetry directly operationalizes Davidson's frontal asymmetry model. Delta MFCCs capture short-term speech dynamics that are commonly associated with affect. These features emphasize signal components that are difficult to learn reliably from limited data.

\paragraph{Implementation Fidelity Dominates in the Low-Sample Regime.} With only $\sim$280 training samples per subject, small implementation mismatches can materially change the effective learning problem and overwhelm the benefits of additional architectural components. We therefore performed targeted consistency checks that preserve the intended objective and inductive bias: (i) ensuring loss--activation compatibility by training on raw logits with \texttt{CrossEntropyLoss} (avoiding redundant softmax operations that distort gradients), and (ii) enforcing a true squeeze-and-excitation bottleneck via a non-trivial reduction ratio (so the module performs channel reweighting rather than a near-identity mapping). These corrections improve methodological validity and, in several cases, recover accuracy without increasing model capacity.

\paragraph{Minimal Complexity.} M3 models add minimal parameters over baselines. The 220-feature audio CNN adds only 40 dimensions. The band power MLP, while larger than EEGNet, operates on preprocessed features rather than raw 15,000-dimensional signals.

\section{Limitations}

Our study has several limitations. First, all experiments use subject-dependent evaluation; cross-subject generalization remains unexplored. Second, the EAV dataset's conversational paradigm may not generalize to spontaneous emotions. Third, we did not explore multimodal fusion, which may further improve results by combining complementary information across modalities. Fourth, our M2 architectures may benefit from more careful hyperparameter tuning on larger validation sets. Finally, the EAV dataset contains only 42 subjects, limiting statistical power for detecting small effects.

\section{Conclusion}

This study systematically investigated attention mechanisms for multimodal emotion recognition on small datasets. Our findings show that for datasets with approximately 280 training samples per subject, domain-appropriate signal processing outperforms sophisticated neural architectures.

Our best results (EEG 67.62\%, +7.62pp; Audio 65.56\%, +3.66pp; Vision 75.30\%, +0.80pp transformer and 72.68\%, +1.28pp CNN) were achieved through simple modifications: frequency-domain features for EEG, delta coefficients for audio and video, and properly pretrained transformers. Meanwhile, our theoretically motivated factorized attention mechanisms (M2) consistently underperformed by 5 to 13 percentage points.

The practical lesson is clear: before adding architectural complexity, researchers should (1) verify baseline implementations for bugs, (2) consider domain-specific feature engineering, and (3) match model capacity to available training data. Future work should evaluate these factorized attention approaches on larger emotion datasets where their sophisticated inductive biases may prove beneficial.

\begingroup
\setlength{\emergencystretch}{2em}
\sloppy
\bibliography{custom}
\endgroup

\onecolumn
\appendix
\section{Appendix: Reproducibility and Additional Details}
\label{sec:appendix}

\small
\setlength{\parindent}{0pt}
\setlength{\parskip}{0.35em}
\setlength{\abovedisplayskip}{0.4em}
\setlength{\belowdisplayskip}{0.4em}

\begin{appendixbox}{A. Reproducibility checklist (what to match)}
\begin{itemize}
\item \textbf{Evaluation protocol:} subject-dependent splits (70/30 train/test) per subject; report mean and std over 42 subjects.
\item \textbf{Input shapes:} EEG $(30\times 500)$ at 100~Hz after downsampling; Audio 5~s at 16~kHz; Video 25 frames sampled at 30~fps.
\item \textbf{Key M3 features:} EEG band power + differential entropy + alpha asymmetry (306 dims); Audio MFCC + $\Delta$MFCC + chroma + mel (220 dims); Vision frame features + temporal deltas.
\item \textbf{Optimization:} AdamW + label smoothing ($\epsilon=0.1$) for tuned baselines.
\end{itemize}
\end{appendixbox}

\begin{appendixbox}{B. Feature definitions used in M3}
\textbf{EEG band power (150 dims).} For each channel $c$ and band $b$, compute Welch PSD $P^{(c)}_{xx}(f)$ and integrate over the band:
\begin{equation}
P_{c,b}=\int_{f_{\mathrm{low}}^b}^{f_{\mathrm{high}}^b} P^{(c)}_{xx}(f)\,df.
\end{equation}

\textbf{EEG differential entropy (150 dims).} For each band-filtered signal (often approximated as Gaussian),
\begin{equation}
\mathrm{DE}_{c,b}=\tfrac{1}{2}\log(2\pi e\sigma^2_{c,b}).
\end{equation}

\textbf{EEG alpha asymmetry (6 dims).} For hemisphere pairs $(i,j)$,
\begin{equation}
\mathrm{Asym}_{ij}=\ln(P^{\alpha}_j)-\ln(P^{\alpha}_i).
\end{equation}

\textbf{Audio (220 dims).} Time-aggregated features: 40 MFCC + 40 $\Delta$MFCC + 12 chroma + 128 mel.

\textbf{Vision (delta features).} For frame embeddings $\mathbf{f}_t$, define $\mathbf{d}_t=\mathbf{f}_{t+1}-\mathbf{f}_t$ and concatenate mean static and mean delta embeddings.
\end{appendixbox}

\begin{appendixbox}{C. EEG frequency bands}
\centering
\begin{tabular}{@{}lll@{}}
\toprule
\textbf{Band} & \textbf{Range (Hz)} & \textbf{Used for} \\
\midrule
$\delta$ & 0.5--4  & band power, DE \\
$\theta$ & 4--8   & band power, DE \\
$\alpha$ & 8--13  & band power, DE, asymmetry \\
$\beta$  & 13--30 & band power, DE \\
$\gamma$ & 30--45 & band power, DE \\
\bottomrule
\end{tabular}
\end{appendixbox}

\begin{appendixbox}{D. Model capacity sanity check (per subject)}
\centering
\begin{tabular}{@{}lrrl@{}}
\toprule
\textbf{Model} & \textbf{Params} & \textbf{Trainable} & \textbf{Accuracy} \\
\midrule
\multicolumn{4}{l}{\textit{EEG}} \\
EEGNet (paper) & 2.6K & 2.6K & 60.0\% \\
M3-EEG v1 (bug-fix) & 2.6K & 2.6K & 57.66\% \\
M3-EEG v2 (features) & $\sim$48K & $\sim$48K & 67.62\% \\
M2 Tri-Stream (small) & 138K & 138K & $\sim$48\% \\
\midrule
\multicolumn{4}{l}{\textit{Audio}} \\
AST (M1) & 87M & $\sim$4K & 58.06\% \\
M3-Audio (Delta) & 87M & $\sim$5K & 65.56\% \\
M2 Dual-Attn. & 87M & $\sim$12M & 49.46\% \\
\midrule
\multicolumn{4}{l}{\textit{Vision}} \\
ViT (M1) & 86M & $\sim$4K & 75.30\% \\
M3-Vision (Delta) & 23M & 23M & 72.68\% \\
M2 Factorized & 86M & $\sim$12M & 69.54\% \\
\bottomrule
\end{tabular}

\medskip
\raggedright
\textbf{Note.} With $\sim$280 training samples per subject, large trainable components can be difficult to constrain and may require stronger regularization or additional data.
\end{appendixbox}

\begin{appendixbox}{E. Augmentations used}
\textbf{EEG:} amplitude scaling, additive Gaussian noise, time shift.

\textbf{Audio:} SpecAugment (time masking and frequency masking).

\textbf{Video:} horizontal flip, color jitter, random erasing.
\end{appendixbox}

\begin{appendixbox}{F. Metrics}
\textbf{Accuracy:}
\begin{equation}
\mathrm{Acc}=\frac{\#\,\text{correct}}{\#\,\text{total}}.
\end{equation}

\textbf{Weighted F1:} compute per-class precision/recall/F1 and average weighted by class support.
\end{appendixbox}

\begin{appendixbox}{G. Short diagnostic summary}
\textbf{Why M2 underperformed:} additional attention modules increase trainable capacity and can perturb pretrained representations; both effects are amplified under limited supervision.

\textbf{Why M3 improved:} hand-crafted features (frequency-domain priors, temporal derivatives) and small implementation fixes provide a stronger inductive bias without substantially increasing effective model complexity.
\end{appendixbox}

\begin{appendixbox}{H. Core equations (used throughout)}
\textbf{Scaled dot-product attention.}
\begin{equation}
\mathrm{Attn}(\mathbf{Q},\mathbf{K},\mathbf{V})=\mathrm{softmax}\!\left(\frac{\mathbf{Q}\mathbf{K}^\top}{\sqrt{d_k}}\right)\mathbf{V}.
\end{equation}

\textbf{Delta features (MFCC).} Using a symmetric window of size $N$,
\begin{equation}
\Delta c_t=\frac{\sum_{n=1}^{N} n\,(c_{t+n}-c_{t-n})}{2\sum_{n=1}^{N} n^2}.
\end{equation}

\textbf{Squeeze-and-excitation (SE).} With channel descriptor $\mathbf{z}=\mathrm{GAP}(\mathbf{x})$ and reduction ratio $r$,
\begin{equation}
\mathbf{s}=\sigma\!\left(\mathbf{W}_2\,\mathrm{ReLU}(\mathbf{W}_1\mathbf{z})\right),\quad \tilde{\mathbf{x}}_c=s_c\mathbf{x}_c.
\end{equation}
\end{appendixbox}

\begin{appendixbox}{I. Model sketches (high-level)}
\textbf{M1 (pretrained transformers).} Fine-tune pretrained AST/ViT with small classifier heads; freeze--unfreeze schedule reduces effective trainable capacity.

\textbf{M2 (factorized attention).} Add modality-specific factorization (EEG: spatial/temporal/asymmetry; Audio: time/frequency; Video: space/time) with skip connections to preserve pretrained pathways.

\textbf{M3 (CNN + features).} Keep architecture simple; inject domain priors via frequency-domain EEG features and temporal derivatives for audio/video.
\end{appendixbox}

\begin{appendixbox}{J. Extended diagnostic analysis (brief)}
\textbf{M2 underperformance:} (i) increased trainable capacity relative to $\sim$280 samples/subject, (ii) sensitivity to perturbations of pretrained features, and (iii) optimization instability from additional fusion/gating components.

\textbf{M3 improvements:} (i) aggressive dimensionality reduction (e.g., 15,000$\rightarrow$306 for EEG), (ii) priors aligned with known signal structure (bands/asymmetry; deltas), and (iii) implementation fidelity (loss--activation compatibility; non-trivial SE bottleneck).
\end{appendixbox}

\end{document}